\documentclass{article}

\pdfoutput=1
\usepackage{arxiv}

\usepackage[utf8]{inputenc} 
\usepackage[T1]{fontenc}    
\usepackage{hyperref}       
\usepackage{url}            
\usepackage{booktabs}       
\usepackage{amsfonts}       
\usepackage{nicefrac}       
\usepackage{microtype}      
\usepackage{lipsum}
\usepackage{graphicx}
\usepackage{caption}

\title{Short-term Demand Forecasting for Online Car-hailing Services using Recurrent Neural Networks}

\author{
  Alireza Nejadettehad\\
  School of Electrical and Computer Engineering\\
  University of Tehran\\
  \texttt{alireza\_ettehad@ut.ac.ir} \\
   \And
  Hamid Mahini \\
  School of Electrical and Computer Engineering\\
  University of Tehran\\
  \texttt{hamid.mahini@ut.ac.ir} \\
   \And
  Behnam Bahrak\\
  School of Electrical and Computer Engineering\\
  University of Tehran\\
  \texttt{bahrak@ut.ac.ir} \\
}

\begin{document}
\maketitle

\begin{abstract}
Short-term traffic flow prediction is one of the crucial issues in intelligent transportation system, which is an important part of smart cities. Accurate predictions can enable both the drivers and the passengers to make better decisions about their travel route, departure time and travel origin selection, which can be helpful in traffic management. Multiple models and algorithms based on time series prediction and machine learning were applied to this issue and achieved acceptable results. Recently, the availability of sufficient data and computational power, motivates us to improve the prediction accuracy via deep-learning approaches. Recurrent neural networks have become one of the most popular methods for time series forecasting, however, due to the variety of these networks, the question that which type is the most appropriate one for this task remains unsolved. In this paper, we use three kinds of recurrent neural networks including simple RNN units, GRU and LSTM neural network to predict short-term traffic flow. The dataset from TAP30 Corporation is used for building the models and comparing RNNs with several well-known models, such as DEMA, LASSO and XGBoost. The results show that all three types of RNNs outperform the others, however, more simple RNNs such as simple recurrent units and GRU perform work better than LSTM in terms of accuracy and training time.
\end{abstract}


\section{Introduction}
Online car-hailing apps have evolved as novel and popular services to provide on-demand transportation service via mobile apps. Comparing with the traditional transportation means such as the subways and buses, the online car-hailing service is much more convenient and flexible for the passengers. Furthermore, by incentivizing private cars owners to provide car-hailing services, it promotes the sharing economy and enlarges the transportation capacities of the cities. Several car-hailing mobile apps have gained great popularities all over the world, such as Uber, Didi, and Lyft. Large number of passengers are served and a significant volume of car-hailing orders are generated routinely every day. For example, TAP30, one of the largest online car-hailing service providers in Iran, handles hundreds of thousands of orders per day all over Iran.

These platforms serve as a coordinator who matches requesting orders from passengers (demand) and vacant registered cars (supply). There exists an abundance of leverages to influence drivers’ and passengers’ preference and behavior, and thus affect both the demand and supply, to maximize profits of the platform or achieve maximum social welfare. Having better understanding of the short-term passenger demand over different spatial zones is of great importance to the platform or the operator, who can incentivize drivers to the zones with more potential passenger demands, and improve the utilization rate of the registered cars. However, in metropolises like Tehran, it is common to see passengers seeking for taxicabs roadside while some taxi drivers are cruising idly on the street. This contradiction reveals the supply-demand disequilibrium with the following two scenarios: Scenario 1, demand exceeds supply, where passengers’ needs would not be met in a timely response. Scenario 2, supply exceeds demand, where drivers would spend overly long time in seeking for passengers. To solve the problem of disequilibrium, an overall prediction for passenger demand in different zones, provides a global distribution of passengers, upon which providers of car-hailing services can adjust prices and dispatch policies of supply dynamically in advance. We define the taxi-demand prediction problem as follows: Given historical taxi demand data in a region $\mathcal{R}$, we want to predict the number of ride requests that will emerge within $\mathcal{R}$ during the next time interval.

Over the past few decades, many data analysis models have been proposed to solve the short-term traffic forecasting problem, including probabilistic models \cite{Yuan2013}, time-series forecasting methods \cite{Li2011}\cite{Moreira-Matias2013} and decision tree based methods\cite{Zhang2017}. Recently approaches based on neural networks gained noticeable attention in studies related to traffic flow prediction\cite{Mukai}\cite{Zhao2016}\cite{Wang2017}. One of the most popular kinds of NNs in this context is Recurrent Neural Networks (RNNs) \cite{Tian2015}\cite{Zhao2017}. Since 2015, when \cite{Tian2015} proposed long-short term memory (LSTM) NNs for traffic flow prediction and showed that LSTMs (due to their excellent ability to memorize long-term dependencies) outperform other methods in this particular context, almost every study that attempted to use RNNs for demand prediction, has utilized LSTMs \cite{Zhao2017}\cite{Xu}\cite{Ke2017}. In this paper, the performance of different types of RNNs are evaluated and compared with some other powerful methods such as eXtreme Gradient Boosting (XGBoost)\cite{Chen2018} and least absolute shrinkage and selection operator (LASSO)\cite{Tibshirani1994} and also with each other. Experimental results demonstrate that RNNs outperform the other methods according to the metrics chosen for comparison; However when it comes to the comparison between RNNs, Simple RNN units and Gated recurrent unit (GRU) defeat LSTM in terms of performance and computational(training) time.

The results obtained from experiments show that the best non-RNN method (XGBoost) reached error rates 3.78 and 40.8\% according to RMSE and MAPE, respectively. However these errors were reduced to 3.22 and 37.42\% by simple RNN units. In addition to the fact that simple RNN units outperformed other non-RNN methods and LSTM, computation time required for simple RNN units is approximately 0.13 and 0.1 the time needed to train XGBoost and LSTM, respectively. Although the experimental results denote that simple RNN units and GRU perform nearly the same, there is a significant difference between their training time and simple RNN units train nearly 13 times faster than GRU.

\section{Related work}
Although there has been many efforts to predict traffic flow using spatiotemporal data; the most related studies to the demand prediction problem shows that the most implemented methods consists of probabilistic models such as Poisson \cite{Yuan2013}, time-series forecasting methods such as auto regression integrated moving average (ARIMA) \cite{Li2011}\cite{Moreira-Matias2013} and neural networks \cite{Mukai}\cite{Zhao2016}\cite{Wang2017}.
Between the time-series forecasting methods, ARIMA is more prevalent because of its performance in short-term forecasting. \cite{Li2011} presented an improved ARIMA-based method to forecast the spatial-temporal distribution of passengers in urban environment. First, urban regions with high demand are detected; then demand in next hour are predicted in those regions using ARIMA and finally, demand is forecasted using an improved ARIMA-based method that uses both time and type of the day. \cite{Moreira-Matias2013} proposes the challenge that ARIMA is not necessarily the best method to forecast demand. They propose an end-to-end framework to predict the number of services that will happen at taxi stands by applying the time-varying Poisson model and ARIMA. Moreover, they used sliding-window ensemble framework to originate a prediction by combining the prediction of each model accuracy. The dataset was generated from 441 vehicles with 63 taxi stands in the city of Porto. \cite{Yuan2013} presented and algorithm based on Poisson model to recommend the most probable points to find passengers for taxi drivers in shortest time. \cite{Davis2016} proposed a multi-level clustering technique to improve the accuracy of linear time-series model fitting, by exploring the correlation between adjacent Geo-hashes.

Recently, the success of deep learning in the fields of computer vision and natural language processing \cite{Lecun2015}\cite{Krizhevsky}, motivated researchers to apply deep learning techniques on traffic prediction problems. \cite{Mukai} is one of the first studies that implemented NNs in order to forecast taxi demand. They have used a multilayer perceptron to achieve this target. \cite{Zhao2016} introduced a new parameter named "Maximum predictability" showed that different predictors (Markov predictor (a probability-based predictive algorithm), the Lempel-Ziv-Welch predictor (a sequence-based predictive algorithm), and the Neural Network predictor (a predictive algorithm that uses machine learning)), perform differently according to the maximum predictability of a region. They showed that considering maximum predictability, in the regions with more random demand pattern, NNs perform better and in the regions with lower randomness in their demand pattern, Markov predictor beats the others. \cite{Wang2017} proposed an end-to-end framework named DeepSD, based on a novel deep neural network structure that automatically discovers the complicated supply-demand patterns in historical order, weather and traffic data, with minimal amount of hand-crafted features.

In 2015 \cite{Tian2015} proposed long-short term memory NNs (LSTMs) for traffic flow prediction and showed that LSTMs (due to their excellent ability to memorize long-term dependencies) perform better in comparison to the other methods in this particular context. Since then, almost every study that used Recurrent neural networks to predict demand, used LSTMs \cite{Zhao2017}\cite{Xu}\cite{Ke2017}. In this paper we are going to compare the performance of different types of RNNs and also evaluate their performance in comparison to some other powerful methods such as XGBoost and LASSO.

\section{Material and Methods}
In this section, first, we explain how we cleaned the dataset and prepared it for modeling. Second, the features used in the models are introduced and finally, three different types of recurrent neural networks that we have used as models are explained in details.

\subsection{Data Processing}
The dataset used in this study is real-world data from TAP30 corporation ride requests from September 1st to December 20th, 2017.
The details of raw data taken from database is shown in Table~\ref{tab1}.
\begin{table}
\centering
\caption{Details of raw data}\label{tab1}
\begin{tabular}{|l|l|}
\hline
{\bfseries Data type} & {\bfseries Description}\\
\hline
Ride Request ID & The unique ID of the ride request\\
Passenger ID & The unique ID of the passenger that made the ride request\\
Timestamp & Timestamp of the ride request\\
Latitude/Longitude & GPS location of origin of the ride request\\
\hline
\end{tabular}
\end{table}
The urban area is partitioned into 16$\times$16 grids uniformly where each grid refers to a region. On the other hand, we consider variables aggregated in a 15 minutes time interval in this paper.
We have removed the ride requests canceled in 5 seconds, because there are not considered to be real demand and potentially are noisy data. And also the ride 
requests that a passenger with his/her unique passenger id has made in a time interval of 15 minutes length are aggregated to become a single request. The number of unique ride requests made, represents the demand. We aggregated the number of unique ride requests for all 256 regions, every 15 minutes.
In order to obtain robust and interpretable results, we decided to consider only the regions that at least 300 ride requests per day on average(nearly 3 ride requests in each time interval on average) had been made in them. After eliminating the regions that does not satisfy our limit, 64 regions were left.

\subsection{Features}
There are 68 main features for the predictive model. Each data point in our final cleaned data has 4 temporal features and 64 spatial features.
\subsubsection{Temporal Features}\label{ss:temporalf}
We have extracted 4 main temporal features from the timestamps of the cleaned raw data.
In order to use the continuous nature of the timeslot feature, first, we converted the timeslot number to triangular format and used its sine and cosine as features. Table~\ref{tab2} includes the temporal features and their description.
\begin{table}
\centering
\caption{Temporal features}\label{tab2}
\begin{tabular}{|l|l|}
\hline
{\bfseries Feature} & {\bfseries Description}\\
\hline
Day of week & The ID of the day of week\\
National holiday & Whether the day is a national holiday or not\\
Timeslot Sineunus & sin(2$\pi \times$timeslot number/96) \\
Timeslot Cosineus & cos(2$\pi \times$timeslot number/96)\\

\hline
\end{tabular}
\end{table}
\subsubsection{Spatial Features}
Since there are correlations between the amount of demand in a region and the other regions, we used the amount of demand in all regions in the previous timeslots as features. For example to predict the demand in timeslot $t+1$ in region number $i$, not only we used the demand in previous timeslots in that region, but also we used the demand in all other regions as features in our models.

\subsection{Methods}
In this section, we briefly describe our selected recurrent neural networks for the aforementioned task, which are Simple RNN, GRU (Gated recurrent unit) and LSTM (Long short term memory).

\subsubsection{Simple RNN}
A recurrent neuron is a special kind of artificial neuron which has a backward connection to the neurons in previous layers. RNNs have internal memory which allows them to operate over sequential data effectively. This feature made the RNNs one of the most popular models for dealing with sequential tasks such as handwriting recognition\cite{Graves2009}, NLP\cite{AlexGraves2013} and time series forecasting\cite{Connor1994}.
\begin{figure}
\centering
\captionsetup{width=.4\linewidth}
\begin{minipage}{.4\textwidth}
\centering
\includegraphics[width=.35\linewidth]{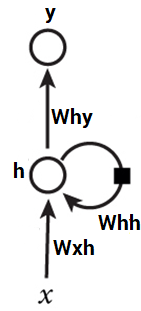}
\caption{A recurrent neural network}
\label{fig:rnn1}
\end{minipage}%
\begin{minipage}{.6\textwidth}
\centering
\includegraphics[width=\linewidth]{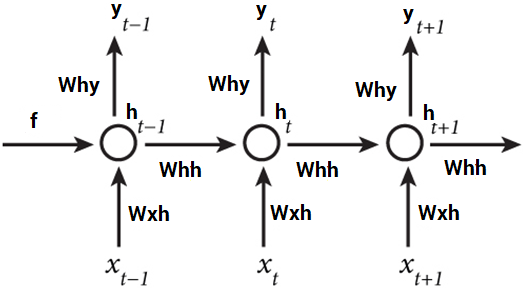}
\caption{An unrolled recurrent neural network}
\label{fig:rnn2}
\end{minipage}
\end{figure}
Figure \ref{fig:rnn1} shows the structure of an RNN and Figure \ref{fig:rnn2} illustrates an unrolled RNN an how it deals with sequential data.
Given a sequence X = \{$x_1$, $x_2$, $x_3$, ..., $x_t$\} as input, RNN computes the hidden state sequence H = \{$h_1$, $h_2$, $h_3$, ..., $h_t$\} and output sequence Y = \{$y_1$, $y_2$, $y_3$, ..., $y_t$\} using Equations \ref{eqn:rnnht} and \ref{eqn:rnnyt}.
\begin{equation}
\label{eqn:rnnht}
h_t = f(W_{hx} x_t + W_{hh} h_{t-1} + b_h)
\end{equation}
\begin{equation}
\label{eqn:rnnyt}
y_t = g(W_{yt} h_t + b_y)
\end{equation}
In Equations \ref{eqn:rnnht} and \ref{eqn:rnnyt} $W_{hx}$, $W_{hh}$ and $W_{yt}$ denote the input-to-hidden, hidden-to-hidden and hidden-to-output weight matrices, respectively. $b_h$ and $b_y$ are hidden layer bias and output layer bias vectors. $f(.)$ and $g(.)$ are the activation functions of the hidden layer and output layer respectively. The hidden state of each time step is passed to the next time step's hidden state.

\subsubsection{Long short term memory}
\begin{figure}
\captionsetup{width=.4\linewidth}
\centering
\begin{minipage}{.5\textwidth}
\centering
\includegraphics[width=\linewidth]{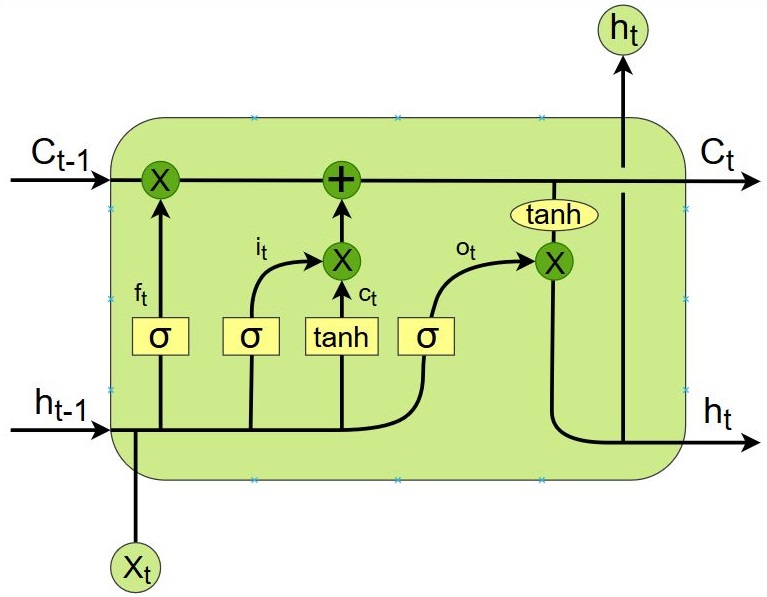}
\caption{An LSTM Cell} \label{fig:lstm}
\end{minipage}%
\begin{minipage}{.5\textwidth}
\centering
\includegraphics[width=\linewidth]{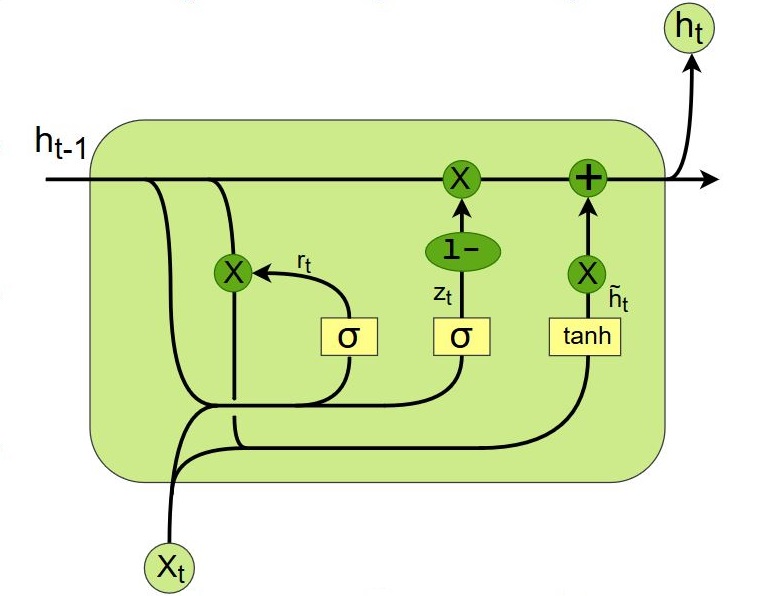}
\caption{A GRU Cell} \label{fig:gru}
\end{minipage}
\end{figure}

Long Short Term Memory networks are a special kind of RNN, capable of learning long-term dependencies. They were introduced by Hochreiter and Schmidhuber (1997)\cite{Hochreiter}, and were refined and popularized by many researchers in different contexts. LSTMs are explicitly designed to avoid the long-term dependency problem. In comparison to simple RNN, LSTM has a more complicated structure and contains three kinds of gates: input gate, forget gate and cell state gate. Figure \ref{fig:lstm} illustrates an LSTM cell. 

Forget gate: After getting the output of previous state, $h(t-1)$, Forget gate helps to take decisions about what must be removed from $h(t-1)$ state and thus keeping only relevant stuff. It is surrounded by a sigmoid function which helps to crush the input between 0 and 1. (Equation \ref{eqn:lstmft}):
\begin{equation}
\label{eqn:lstmft}
f_t = \sigma(W_f . [h_{t-1}, x_t] + b_f)
\end{equation}
Input Gate: In the input gate, we decide to add new stuff from the present input to our present cell state scaled by how much we wish to add them. Sigmoid layer decides which values to be updated and $tanh$ layer creates a vector for new candidates to added to present cell state. (Equations \ref{eqn:lstmit} and \ref{eqn:lstmchat}):
\begin{equation}
\label{eqn:lstmit}
i_t = \sigma(W_i . [h_{t-1}, x_t] + b_i)
\end{equation}
\begin{equation}
\label{eqn:lstmchat}
\hat{C}_t = tanh(W_C. [h_{t-1}, x_t] + b_C)
\end{equation}
Then the cell state is calculated by Equation \ref{eqn:lstmc}:
\begin{equation}
\label{eqn:lstmc}
C_t = f_t * C_{t-1} + i_t * \hat{C}_t
\end{equation}
Output Gate: Finally the sigmoid function decides what to output from the cell state as shown in Equation \ref{eqn:lstmot}. We multiply the input with “tanh” to crush the values between (-1) and 1, then multiply it with the output of sigmoid function so that we only output what we want to. (Equations \ref{eqn:lstmot} and \ref{eqn:lstmht})
\begin{equation}
\label{eqn:lstmot}
o_t = \sigma(W_o . [h_{t-1}, x_t] + b_o)
\end{equation}
\begin{equation}
\label{eqn:lstmht}
h_t = o_t * tanh(C_t)
\end{equation}

\subsubsection{Gated recurrent unit}
GRU was proposed by Cho et al. in 2014\cite{Cho2014}. It is similar to LSTM in structure but simpler to compute and implement. The difference between a GRU cell and an LSTM cell is in the gating mechanism. It combines the forget and input gates into a single update gate. It also merges the cell state and the hidden state. The function of reset gate is similar to forget gate of LSTM. Since the structure of GRU is very similar to LSTM, we will not get into the detailed formula. The structure of a GRU cell is shown in Figure \ref{fig:gru}.

\subsection{Methods for Comparison}\label{ss:compmeths}
We compared the results obtained from recurrent neural networks with a tree-based regression method (XGBoost), one linear regression method (LASSO) and one moving average time series forecasting method (DEMA). We have tuned the parameters for all these methods, then reported the results.
Since these methods are not able to process sequentially formed data, demand intensity for 4 previous timeslots (the sequence length chosen for RNNs) were fed to them as features.
\subsubsection{DEMA}
Double exponential moving average is a well-known method for time series forecasting problems. It attempts to remove the inherent lag associated to Moving Averages by placing more weight on recent values. The name suggests this is achieved by applying a double exponential smoothing which is not the case. The name double comes from the fact that the value of an EMA (Exponential Moving Average) is doubled. To keep it in line with the actual data and to remove the lag the value "EMA of EMA" is subtracted from the previously doubled EMA.
\begin{equation}
\label{eqn:dema}
DEMA = 2\times EMA - EMA(EMA)
\end{equation}
\subsubsection{LASSO}
Least absolute shrinkage and selection operator(LASSO) is a linear model that estimates sparse coefficients. It usually produces better prediction result than simple linear regression. We use the LASSO implementation from the scikit-learn library.\cite{Tibshirani1994}
\subsubsection{XGBoost}
eXtreme Gradient Boosting(XGBoost) is a powerful ensemble boosting tree based method and is widely used in data mining applications both for classification and regression problems. We use the XGBoost implementation from XGBoost python package.\cite{Chen2018}

\section{Results}
In this section, we declare our RNNs' specifications and introduce the metrics that evaluations are performed based on them. Then, we evaluate different RNN models on our dataset and see how well they can predict the requests in the future. In addition, we compare our model with 3 other baselines and show that RNNs outperform all.

\subsection{Experimental Setup}
Our dataset is obtained from TAP30 Co. ride requests in Tehran from September 1st to December 20th, 2017. We used the first prior 80 days to train the models and last 30 days for validation. All three kinds of recurrent neural networks (Simple RNN, GRU, LSTM) were implemented in Keras API built on top of Tensorflow. Although recurrent neural networks can accept sequences with any length as input, because of the nature of our problem we had to choose a constant sequence length. Due to the constrained computational power we had, we used every hour data as a sequence. Because the time interval for each data point is 15 minutes, each sequence consists of four data points. Since the data contains records for 110 days, the shape of data would be (110*24, 4, 68). Table~\ref{tab3} includes the list of parameters used in the experiment for all three types of RNNs.
\begin{table}
\centering
\caption{Experimental Parameters}\label{tab3}
\begin{tabular}{|l|l|}
\hline
Data of each sequence & 1 hour data\\
Time-step length & 15 mins\\
Sequence length & 4\\
Number of regions & 64\\
Number of features & 68\\
Number of hidden layers & 2\\
Number of neurons in each hidden layer & 1500-2000\\
Activation function of hidden recurrent layers & tanh\\
Loss function & Mean squared error\\
\hline
\end{tabular}
\end{table}
\subsection{Evaluation metrics}
We use root mean absolute error (RMSE) and mean absolute percentage error (MAPE) to evaluate the models. These metrics are defined as follows:
\begin{equation}
\label{eqn:rmse}
RMSE = \sqrt{\frac{1}{n}\sum_{i=1}^{n} (y_{t+1}^i - \hat{y}_{t+1}^i)^2}
\end{equation}
\begin{equation}
\label{eqn:mape}
MAPE = \frac{1}{n}\sum_{i=1}^{n} \frac{|y_{t+1}^i - \hat{y}_{t+1}^i|}{y_{t+1}^i}
\end{equation}
Where $y_{t+1}^i$ and $\hat{y}_{t+1}^i$ mean the real and prediction value for demand in region $i$ for time interval $t+1$ and $n$ denotes total number of samples.

\subsection{Experimental Results}
First we report the performance of RNNs (RMSE and MAPE) over the entire city (all selected regions) and then we report the errors on each category of regions.
\subsubsection{Performance over the entire city}
\begin{figure}
\captionsetup{width=.4\linewidth}
\centering
\begin{minipage}{.5\textwidth}
\centering
\includegraphics[width=\linewidth]{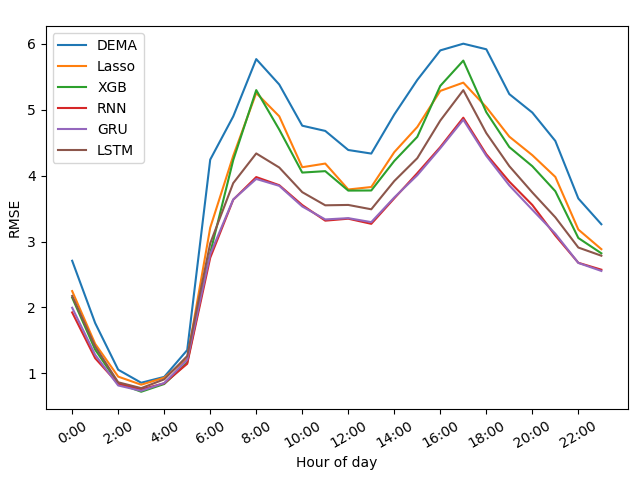}
\caption{Prediction performance at different hours according to RMSE}
\label{fig:rmse-hourly}
\end{minipage}%
\begin{minipage}{.5\textwidth}
\centering
\includegraphics[width=\linewidth]{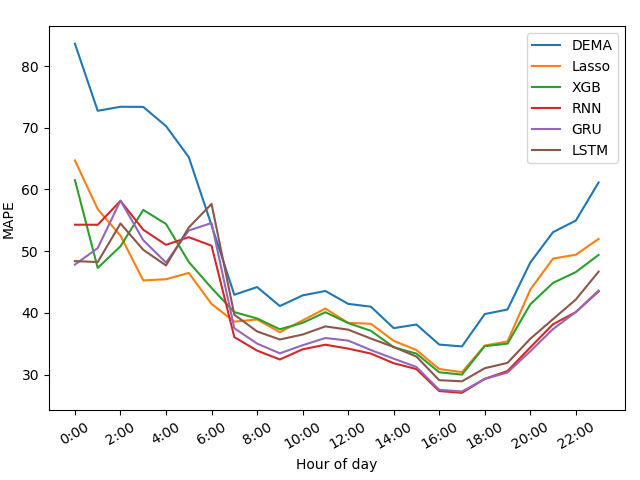}
\caption{Prediction performance at different hours according to MAPE}
\label{fig:mape-hourly}
\end{minipage}
\end{figure}

To evaluate the prediction performance over the entire city which includes 64 regions, we compare the performance of RNNs with other methods described in \ref{ss:compmeths} in terms of RMSE and MAPE from Equations \ref{eqn:rmse} and \ref{eqn:mape}. 
We report the RMSE and MAPE over the entire city during daily hours in Figures \ref{fig:rmse-hourly} and \ref{fig:mape-hourly}.

As it can be seen in Figures \ref{fig:rmse-hourly} and \ref{fig:mape-hourly}, all methods share common patterns through both metrics. For instance, they reach their minimum values at about 3am and maximum values at about 7pm. All three kinds of RNNs show better performance than the other methods, but between them, RNN and GRU have nearly the same error values during the day and are better than LSTM with a considerable difference. There is a haphazard pattern between hours 12:00am and 6:00am in Figure \ref{fig:mape-hourly}. According to the Equation \ref{eqn:mape}, MAPE is a very sensitive metric and depends on the real value's range. Since the amount of ride requests through these hours are extremely low, this metric fails to have a specific pattern during these hours. Predicting demand intensity during rush hours (about 8am and 5pm) is considered more crucial than the other times. According error rates both RMSE and MAPE, it can be observed that RNNs demonstrate considerably better performance in comparison to the others. 
\begin{table}
\centering
\caption{Errors over the entire city}\label{tab4}
\begin{tabular}{|p{3cm}|c|c|c|}
\hline
{\bfseries Method} & {\bfseries RMSE} & {\bfseries MAPE (\%)} & {\bfseries Training time}\\
\hline
DEMA & 4.37 & 48.54 & -\\
LASSO & 3.87 & 41.42 & 4 mins/37 secs\\
XGBoost & 3.78 & 40.80 & 120 mins/53 secs\\
LSTM & 3.46 & 39.04 & 146 mins/43 secs\\
Simple RNN & 3.22 & 37.42 & 16 mins/40 secs\\
GRU & 3.21 & 37.50 & 119 mins/19 secs\\
\hline
\end{tabular}
\end{table}
Table \ref{tab4} shows the detailed values of errors over the entire city for each method. Training was performed on a core-i7-7700HQ CPU with 16 GBs of RAM.

\subsubsection{Performance over categorized regions}
\begin{figure}
\captionsetup{width=.4\linewidth}
\centering
\begin{minipage}{.5\textwidth}
\centering
\includegraphics[width=\linewidth]{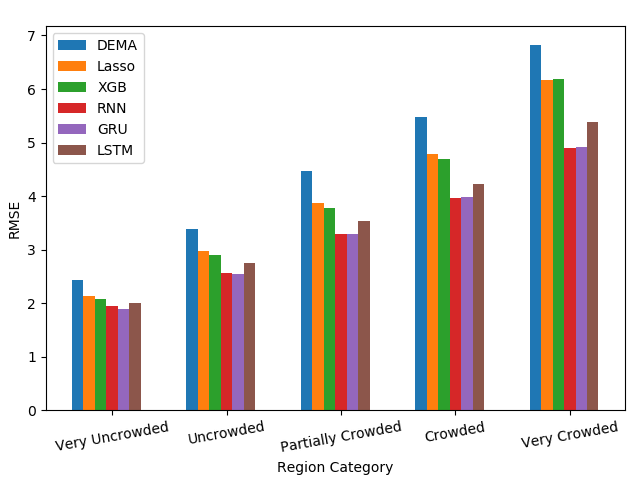}
\caption{Prediction performance in different region categories according to RMSE}
\label{fig:rmse-reg-cat}
\end{minipage}%
\begin{minipage}{.5\textwidth}
\centering
\includegraphics[width=\linewidth]{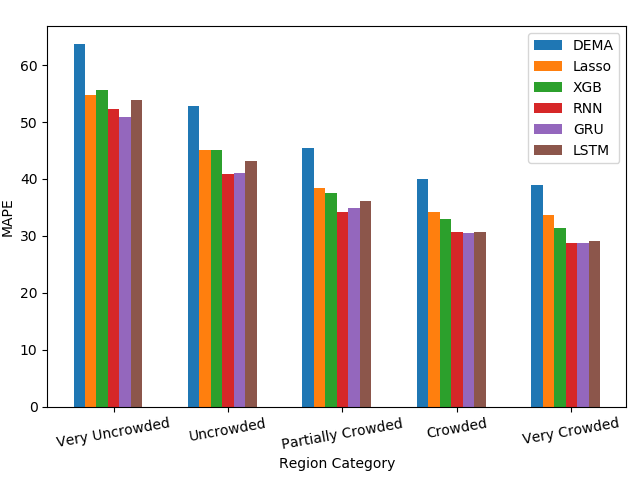}
\caption{Prediction performance in different region categories according to MAPE}
\label{fig:mape-reg-cat}
\end{minipage}
\end{figure}

We have categorized 64 regions in Tehran, to 5 distinct categories. The regions with average ride requests per day greater than 1600, are categorized as very crowded regions and the regions with average ride requests per day less than 400, are categorized as very uncrowded regions and the other 3 categories are placed between these 2 categories. Figures \ref{fig:rmse-reg-cat} and \ref{fig:mape-reg-cat} illustrate the performance in terms of RMSE and MAPE respectively over these 5 categories. As we move from the very uncrowded regions to very crowded ones, since the real value of demand gets greater, the range for RMSE gets greater and the range for MAPE becomes less. But over all 5 categories, RNNs show a better performance. Especially simple RNN and GRU are the best models.

\section{Conclusion}
In this paper different types of recurrent neural networks were implemented and used in order to forecast short-term demand in different regions on an online car-hailing company's data. We compared the performance of prediction between three types of RNNs including simple RNN, GRU and LSTM with tree based models (XGBoost and Random forest), a very powerful linear regression model (LASSO) and time series forecasting models based on moving averages (SMA, DEMA). The results indicated that all three types of RNNs outperformed the other methods but the simple RNN and GRU showed the best results between RNNs. Compared to the best non-RNN method (XGBoost), GRU and Simple RNN reduced RMSE about 15\% and reduced MAPE nearly 8\%.
Since the nature of the demand prediction problem for traffic flow is a short-term history dependent kind, more simple types of RNN's performed better than long-short term memory networks (LSTM). Not only LSTM networks' performance is worse than other RNNs, but also it takes more time for training due to the complexity of these networks.

\section{Acknowledgments}
This research is financially supported by TAP30 Co. and also the authors are grateful to TAP30 Co. for providing sample data.

%
%
\bibliographystyle{elsarticle-num}

\end{document}